\documentclass[10pt,twocolumn,letterpaper]{article}

\usepackage[pagenumbers]{cvpr} 

\usepackage{amsmath}
\usepackage{graphicx}
\usepackage[colorlinks=true]{hyperref}
\usepackage{tabularx}
\usepackage{adjustbox}
\usepackage{algorithm}
\usepackage{algpseudocode}
\usepackage{color, colortbl}
\usepackage{textcomp}
\definecolor{Gray}{gray}{0.9}
\definecolor{LightCyan}{rgb}{0.88,1,1}
\definecolor{cherryblossompink}{rgb}{1.0, 0.72, 0.77}
\definecolor{carrotorange}{rgb}{0.93, 0.57, 0.13}
\definecolor{amber}{rgb}{1.0, 0.75, 0.0}
\definecolor{bleudefrance}{rgb}{0.19, 0.55, 0.91}
\definecolor{blizzardblue}{rgb}{0.67, 0.9, 0.93}
\definecolor{babyblue}{rgb}{0.54, 0.81, 0.94}
\usepackage{multirow}
\usepackage{makecell}
\usepackage{float}

\title{Domain Adaptable Fine-Tune Distillation Framework For Advancing Farm Surveillance}

\author{Raza Imam$^1$\thanks{This work is done as the part of the environmental project at Fujairah Research Center, Fujairah Environment Authority, UAE.}\quad Muhammad Huzaifa$^1$\quad Nabil Mansour$^2$\quad Shaher Bano Mirza$^2$ \quad Fouad Lamghari$^2$ \\
$^1$Mohamed bin Zayed University of Artificial Intelligence (MBZUAI), UAE\\
$^2$Fujairah Research Center, UAE\\
{\tt\small raza.imam@mbzuai.ac.ae; muhammad.huzaifa@mbzuai.ae}
}

\begin{document}
\maketitle

\begin{abstract}
In this study, we propose an automated framework for camel farm monitoring, introducing two key contributions: the Unified Auto-Annotation framework and the Fine-Tune Distillation framework. The Unified Auto-Annotation approach combines two models, GroundingDINO (GD), and Segment-Anything-Model (SAM), to automatically annotate raw datasets extracted from surveillance videos. Building upon this foundation, the Fine-Tune Distillation framework conducts fine-tuning of student models using the auto-annotated dataset. This process involves transferring knowledge from a large teacher model to a student model, resembling a variant of Knowledge Distillation. The Fine-Tune Distillation framework aims to be adaptable to specific use cases, enabling the transfer of knowledge from the large models to the small models, making it suitable for domain-specific applications.
By leveraging our raw dataset collected from Al-Marmoom Camel Farm in Dubai, UAE, and a pre-trained teacher model, GroundingDINO, the Fine-Tune Distillation framework produces a lightweight deployable model, YOLOv8. This framework demonstrates high performance and computational efficiency, facilitating efficient real-time object detection. Our code is available at \href{https://github.com/Razaimam45/Fine-Tune-Distillation}{https://github.com/Razaimam45/Fine-Tune-Distillation}
\end{abstract}


\section{Introduction}
Computer vision is transforming animal farm management by automating tasks, monitoring animal health, and enhancing security. Vision systems analyze visual cues like body condition and behavior to detect illness early \cite{koger2023quantifying}. Image recognition algorithms monitor feed intake, ensuring optimal nutrition \cite{tassinari2021computer}. AI-based surveillance enhances farm security by detecting unauthorized access \cite{abd2020review}. In camel farming, computer vision aids in reproductive management, improving breeding success \cite{abd2020review}. Visual data analysis provides insights for informed decision-making in herd management and resource allocation \cite{barnard2016quick}.
Recent advances in AI, particularly in large multimodal models like Segment Anything Model \cite{kirillov2023segment}, GPT-4 \cite{openai2023gpt4}, and BLIP2 \cite{li2023blip2}, have expanded the capabilities of computer vision significantly. However, these models have limitations, including computational demands and access constraints \cite{jia2022visual}. Knowledge distillation techniques offer a way to leverage their power in practical applications \cite{gou2021knowledge}.

\begin{figure}[b]
    \centering
    \includegraphics[width=0.4\textwidth]{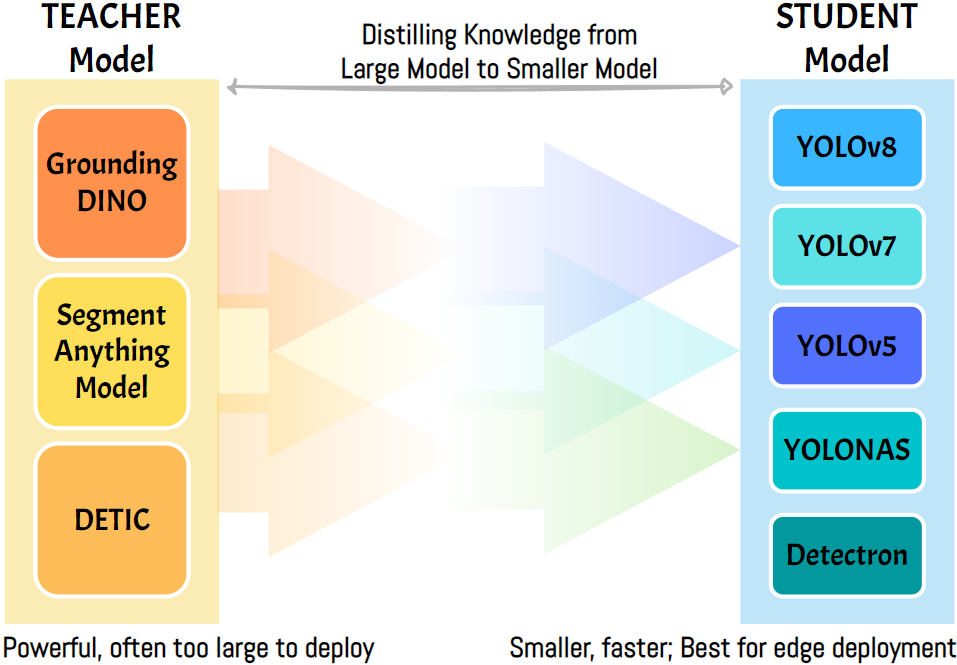}
    \caption{Overview of the Knowledge Transfer in Fine-Tune Distillation}
    \label{fig:overview_ft-distill}
\end{figure}

Foundation models, while versatile and knowledgeable, are impractical for real-time deployment due to size and computational demands \cite{wang2023review}. Fine-tuning them for specific tasks like camel farming is cumbersome. To enhance the taming process, detecting neck ropes, leg ropes, and masks on camels is crucial. This enables safe and effective taming techniques by ensuring proper positioning and tension of ropes and mask usage. Continuous monitoring for abnormalities during taming is essential for camel welfare and farm operations, thus presenting a requirement for a framework to achieve these objectives.
This work aims to develop an automated camel monitoring system using State-Of-The-Art computer vision algorithms. We propose two frameworks: Unified Auto-annotation and Fine-Tune Distillation. The Unified Auto-annotation framework, comprising GroundingDINO and SAM \cite{liu2023grounding} \cite{kirillov2023segment}, generates annotations for object detection and instance segmentation. The Fine-Tune Distillation framework builds upon this, utilizing knowledge distillation to fine-tune the student model (YOLOv8 \cite{Yolov8AimAssist}). This process enhances object detection while reducing computational requirements.
The system focuses on identifying camels, leg/neck ropes, poles, and masks in CCTV footage from a camel farm. Real-time monitoring ensures camel well-being and safety. The integration of these methods enhances camel farm surveillance, improving monitoring and management practices.
\begin{enumerate}
\itemsep-0.05em
    \item Proposed a transferable unified Auto-Annotation approach using GroundingDINO and Segment-Anything-Model (SAM) for automatic annotations of camel farm surveillance videos.
    \item Introduced the Fine-Tune Distillation framework, building on the Auto-Annotation approach, for knowledge distillation to create an efficient detection model.
    \item Evaluated the performance of the fine-tuned student model, YOLOv8s, on a camel farm dataset, demonstrating its effectiveness in real-time surveillance streams.
    \item Conducted a comparative analysis of different student model configurations in terms of computational parameters for potential deployment in real-time camel farm settings. 
\end{enumerate}

\section{Literature Review}
Recent studies have addressed challenges in camel detection, monitoring, and biometric measurement using image processing, computer vision, and IoT \cite{falomir2011describing}. These works combine machine learning, deep learning, and data mining to enhance road safety, conservation, and livestock management.
In the context of camel detection and safety, Madi et al. \cite{madi2023camel} developed a framework that employs machine learning and computer vision to mitigate camel-vehicle accidents, reducing collisions in the Middle East. Alnujaidi et al. \cite{alnujaidi2023computer, alnujaidi2023spot} tested deep learning-based models, with YOLOv8 showing superior performance in camel detection on Saudi Arabian roads.
For camel biometrics, Khojastehkey et al. \cite{khojastehkey2019biometric} utilized machine vision to accurately estimate one-humped camel body dimensions, offering an efficient alternative to manual assessments.

Advancements in image analysis and Deep Neural Networks benefit livestock in general. Achour et al. \cite{achour2020image} developed a non-invasive system for individual dairy cow identification and feeding behavior monitoring, achieving high scores. Shao et al. \cite{shao2020cattle} proposed a cattle detection and counting system using CNNs and UAVs for grazing cattle management.
To address the challenge of real-time monitoring of body weight and daily gain in beef cattle, Cominotte et al. \cite{cominotte2020automated} developed an automated computer vision system. They used biometric body measurements from three-dimensional images as explanatory variables for predictive approaches, with the Artificial Neural Network approach outperforming others. Xu et al. \cite{xu2020automated} demonstrated the application of Mask R-CNN for automated cattle counting in aerial imagery, surpassing existing algorithms in accuracy.
These image analysis and deep learning advances enhance livestock farming, optimizing productivity, health monitoring, cattle management, and accurate counting \cite{madi2023camel}-\cite{xu2020automated}.
Table \ref{tab:related_works} also summarizes the discussed literatures in a compact manner. 

\begin{figure}[b]
    \centering
    \includegraphics[width=0.45\textwidth]{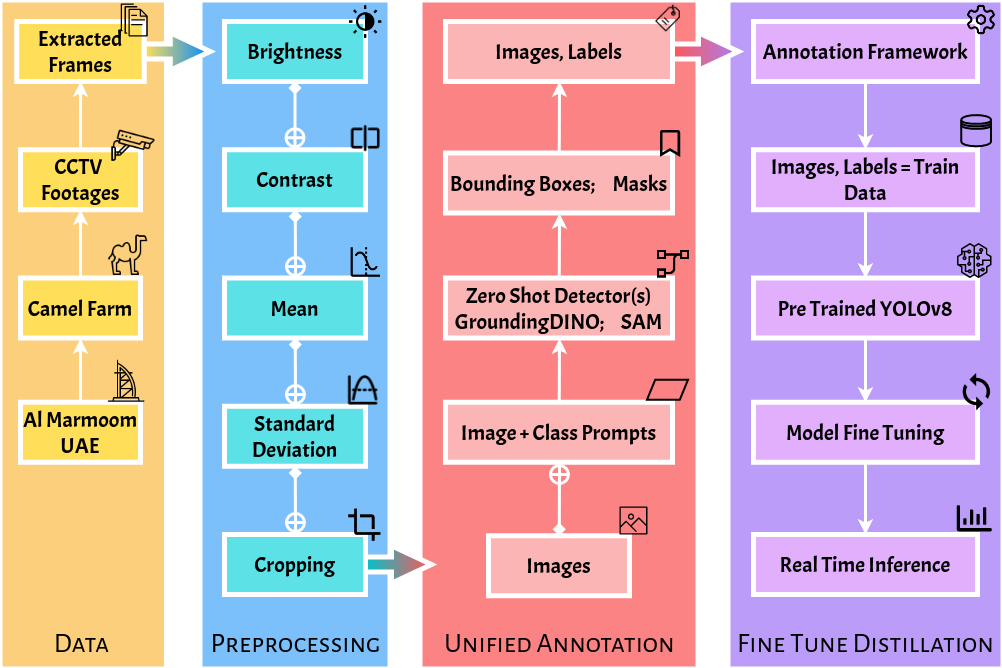}
    \caption{An overview of the comprehensive research design implemented in this work}
    \label{fig:research_design}
\end{figure}

\begin{figure}[b]
    \centering
    \includegraphics[width=0.45\textwidth]{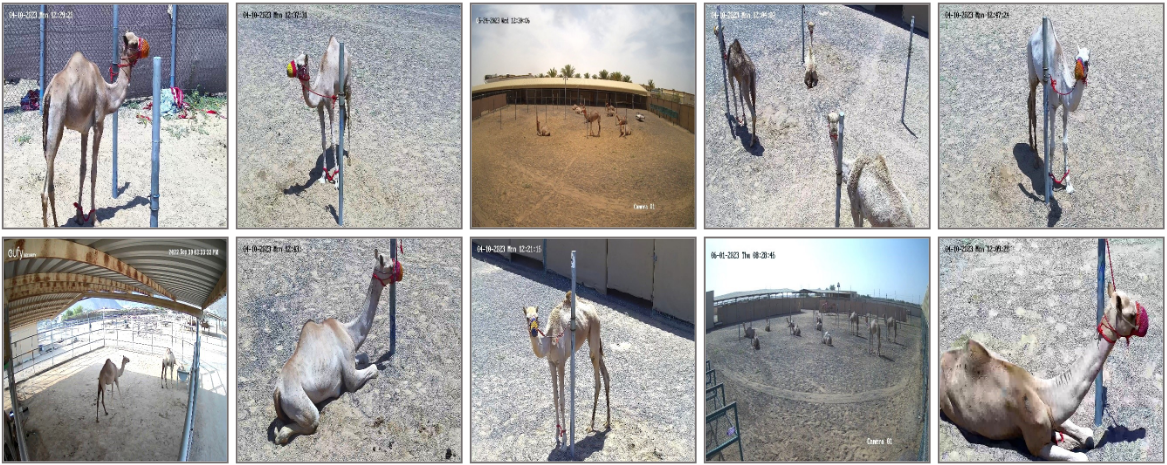}
    \caption{Sample examples of our dataset following the the preprocessed phase}
    \label{fig:dataset_eg}
\end{figure}

\section{Research Design and Preprocessing}
This work aims to develop an automated camel monitoring and welfare assessment system in a farm environment using SOTA vision approaches. The framework detects camels, leg/neck ropes, poles, and masks in real-time for their well-being and safety. This design is adaptable to other livestock farms, automating detection in various settings. The research design consists of Data Collection, Annotation, Preprocessing, Modeling, and Inference stages, forming a systematic approach. The adapted research design is illustrated in Figure \ref{fig:research_design}.

\textbf{Data Collection.} In the initial Data Collection stage, we have two sub-stages: recording surveillance footage and extracting frames. Surveillance footage is gathered from CCTV cameras and monitoring devices at Al-Marmoom Camel farm in Dubai, UAE, with a total of 53 camels. Recordings are made at different times of the day, focusing on taming activities during April, May, and early June 2023. In the frame extraction sub-stage, we extract every $10^{\text{th}}$ frame from the recordings, forming the basis for subsequent analysis. This systematic approach ensures the availability of authentic and high-quality data for reliable research \ref{fig:research_design}.

\begin{figure}[t]
    \centering
    \includegraphics[width=0.4\textwidth]{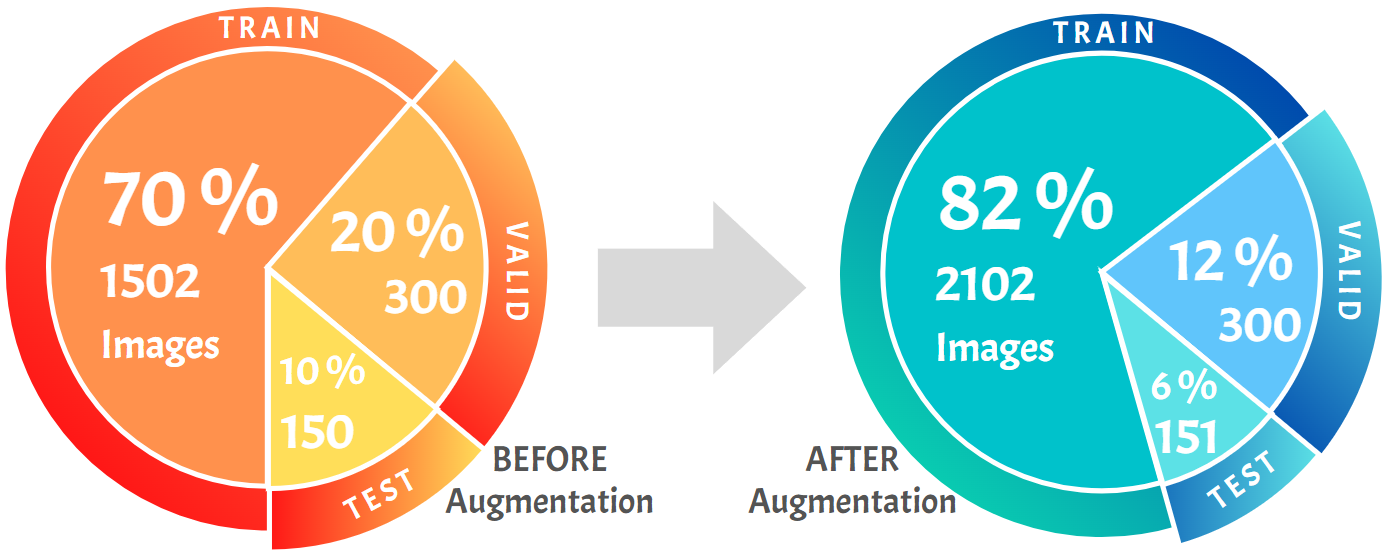}
    \caption{Data distribution before and after augmentation stage}
    \label{fig:data_dist}
\end{figure}


\textbf{Annotation.} Following the frame extraction sub-stage, we have a total of 1502 raw images. The subsequent annotation stage involves marking and labeling specific regions or objects of interest in these images. To automate this process, we employed our proposed transferable and unified framework, incorporating state-of-the-art large vision models like GroundingDINO (GD) and Segment-Anything-Model (SAM) \cite{liu2023grounding, kirillov2023segment}. This automated annotation results in labeled bounding boxes indicating object presence and location within the images. These annotations serve as ground truth for model training and evaluation. Further technical details regarding this methodology is explained in Section \ref{sec:auto_annotation}. 

\textbf{Preprocessing and Augmentation.} After annotation, our dataset underwent preprocessing to enhance image quality, including resizing, brightness/contrast adjustment, optional normalization, and noise reduction \cite{wang2019review}, \cite{fan2019brief} (Table \ref{tab:preprocessing}). This step benefits specific types of analysis or model training \cite{fan2019brief}. We divided the 1502 preprocessed images into three sets: training (70\%, 1051 images), validation (20\%, 300 images), and testing (10\%, 151 images). In the augmentation stage, we applied various transformations to augment the training dataset, effectively doubling its size. This strategy exposed models to diverse samples and variations, aiding generalization to unseen data. Techniques included cropping with 0-35\% zoom \cite{xu2023comprehensive} and grayscale conversion for 20\% of training images (Figure \ref{fig:dataset_eg}). Grayscale augmentation enhances model adaptability to color variations and emphasizes other visual cues \cite{shorten2019survey}, \cite{newey2018shadow}. Our dataset now consists of 2553 images, distributed as follows: training (82\%, ~2102 images), validation (12\%, ~300 images), and testing (6\%, ~151 images) (Figure \ref{fig:data_dist}).

\begin{figure}[t]
\centering
\includegraphics[width=0.45\textwidth]{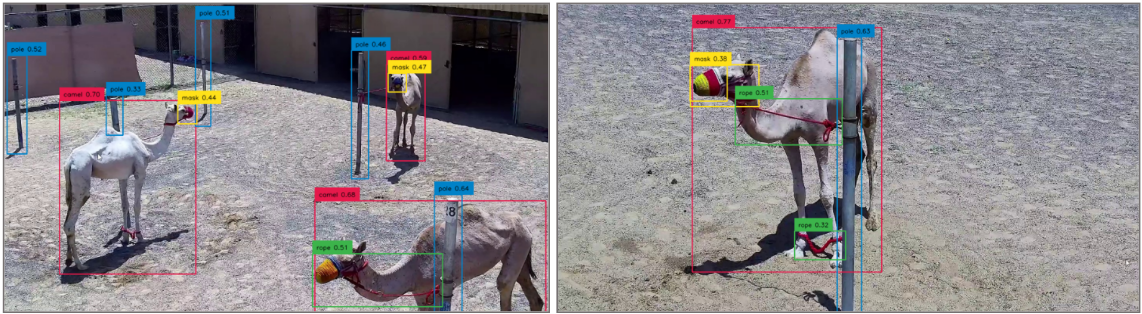}
\caption{Zero-Shot Inference on our dataset images utilizing GroundingDINO with the class prompts "camel", "rope", "mask", and "pole" as the four classes of interest in the context of the taming process. (\textcolor{red}{red} BB (Bounding Box) denotes class "camel", \textcolor{green}{green} denotes "rope", \textcolor{yellow}{yellow} denotes "mask", and \textcolor{blue}{blue} BB denotes "pole")}
\label{fig:zero_infer_use_case}
\end{figure}

\begin{figure*}[t]
    \centering
    \includegraphics[width=0.8\textwidth]{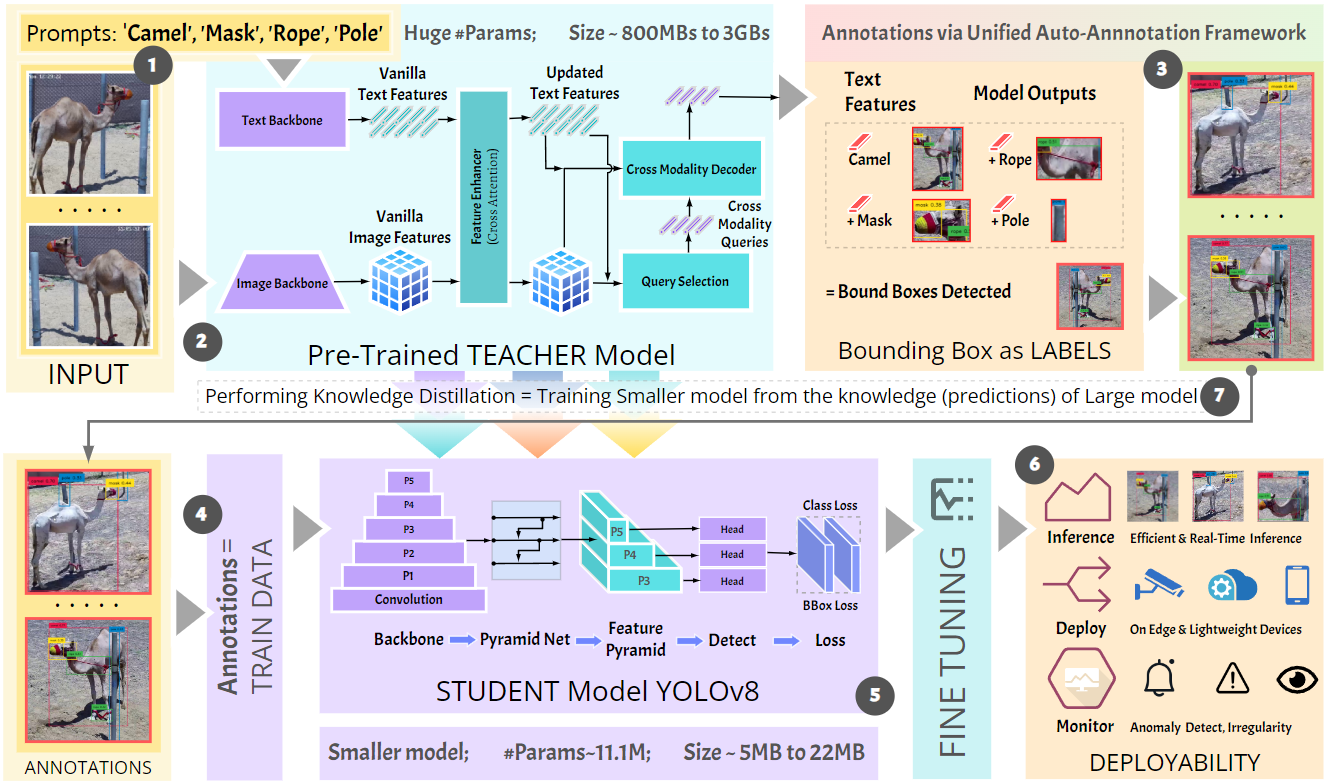}
    \caption{Workflow of the Fine-Tune Distillation framework. (1) Images and class prompts are inputted to Fine-Tune Distillation framework that utilizes (2) Unified Auto-Annotation by performing zero-shot predictions to output (3) bounding boxes respective to prompted classes. These bounding boxes then act as (4) training labels for fine-tuning a smaller and (5) lightweight (student) model. (6) The resultant student model is deployed on edge devices for real-time inference, (7) representing a variation of Knowledge Distillation from Teacher to Student model.}
    \label{fig:method3_distill}
\end{figure*}

\section{Proposed Method}
\label{sec:zero_infer}

Open-set object detection, as demonstrated by GroundingDINO \cite{liu2023grounding}, combines DINO \cite{zhang2022dino} and GLIP \cite{li2022grounded} to detect known and unknown objects, facilitating recognition of novel objects. Figure \ref{fig:zero_infer_generic_camel} illustrates zero-shot inference using GroundingDINO on our camel dataset. This model is ideal for prompt-based zero-shot inference tasks (Figure \ref{fig:method1_zero_infer}). GroundingDINO's integration of text information and grounded pre-training enables prompt-based inference on camel farm surveillance, eliminating the need for retraining. By adjusting prompts, we efficiently identify objects of interest, saving time and resources. Its Transformer-based fusion of cross-modal features processes image and text data, enhancing accuracy and generalization in zero-shot detection tailored to camel taming (Figure \ref{fig:zero_infer_use_case}) \cite{li2022dn, li2023mask, liu2022dab}.

\subsection{Unified Auto-Annotation Framework}
\label{sec:auto_annotation}
Image annotation for tasks like object detection or instance segmentation can be a laborious and costly endeavor. We propose a Unified Auto-Annotation Framework that automates image annotation using GroundingDINO (GD) and Segment-Anything-Model (SAM) \cite{liu2023grounding, kirillov2023segment}. This framework streamlines annotation, eliminating manual effort, and enhances efficiency in dataset preparation for object detection and segmentation tasks. It offers a scalable approach for downstream tasks, ensuring seamless integration with annotated data.
The auto-annotation framework employs GroundingDINO for zero-shot detection and SAM for converting bounding boxes into segmentation masks, detecting objects like "camel," "rope," "mask," and "pole." This two-step process provides both bounding box annotations and detailed masks, as depicted in Figure \ref{fig:method2_auto_annotation}. Annotations are then transformed into various formats (.json, .xml, .txt, .csv) to serve as labels, ensuring adaptability to different tasks and domains as shown in Algorithm \ref{alg:annotation-process}. This data-agnostic framework automates annotation using GroundingDINO and SAM, delivering high-quality labels suitable for diverse use-cases \cite{birmingham2022multi}.

\subsection{Fine-Tune Distillation in Real-time monitoring}
While large vision models like GroundingDINO and SAM offer impressive zero-shot capabilities, their real-time use and broad applicability can be limited. These models possess extensive but generalized knowledge, making them less suitable for specialized domains and resource-constrained applications. To address this, knowledge distillation is employed, where a complex \textit{teacher} model (e.g., GroundingDINO) transfers its knowledge to a smaller \textit{student} model (e.g., YOLOv8). This training process ensures that the student model captures the teacher's insights by matching soft targets (probability distributions), rather than replicating hard predictions. Knowledge distillation allows for efficient deployment in scenarios with computational constraints or specialized requirements.

\noindent \textbf{Fine-Tune Distillation.} 
We introduce Fine-Tune Distillation, an extension of the Unified Auto-annotation framework, inspired by Knowledge Distillation, aimed at enabling efficient real-time object detection. This approach allows for knowledge transfer from a teacher model, in our case, GroundingDINO, which is pretrained on a vast dataset like ImageNet and demands significant computational resources \cite{zhang2022autodistill}. Fine-Tune Distillation leverages GroundingDINO as the teacher model and fine-tunes a state-of-the-art student model, YOLOv8 \cite{yolov8terven2023comprehensive}, on the annotated dataset. However, it is adaptable to different use cases, offering flexibility to choose teacher models like DETIC \cite{zhou2022detecting}, OWL-ViT \cite{minderer2022simple}, or GLIPv2 \cite{zhang2022glipv2}, and student models such as YOLOv7 \cite{wang2023yolov7}, YOLOv5 \cite{Jocher_YOLOv5_by_Ultralytics_2020}, YOLO-NAS \cite{YOLONAS_supergradients}, DETR \cite{carion2020endtoend}, Detectron2 \cite{cheng2020panopticdeeplab}, among others.
The process starts with annotating images via zero-shot inference using the teacher model, GroundingDINO, followed by fine-tuning the student model, YOLOv8, using the resulting annotated dataset. The objective is to align the student model's probability distributions with those generated by GroundingDINO. This approach streamlines real-time object detection while ensuring adaptability to diverse domains. The knowledge transfer in Fine-Tune Distillation is depicted in Figure \ref{fig:overview_ft-distill}.

\noindent \textbf{Working of Fine-Tune Distillation.}
To deploy a model using Fine-Tune Distillation as depicted in Figure \ref{fig:method3_distill}, several steps are involved. First, prepare a dataset tailored to the specific use case, such as camel farm monitoring, including relevant classes like "camel," "rope," "mask," and "pole."
Next, configure inputs for Fine-Tune Distillation, utilizing a teacher model like GroundingDINO. The teacher model, guided by prompts, automatically annotates input images. These auto-labeled images are then used for training the student model.
The student model, based on an architecture like YOLOv8, is fine-tuned using the annotations generated by GroundingDINO. This process results in a smaller model capable of accurately identifying target classes.
Once fine-tuning is complete, deploy the model for real-time applications, offering high FPS on various devices, suitable for diverse scenarios. The pseudocode of the proposed method is given as Algorithm \ref{alg:fine-tuning}.

\section{Experiments}
\subsection{Setup}
Our experiments were conducted on an Nvidia Quadro RTX 6000 GPU with 24 GB of memory. Within our Fine-Tune Distillation framework, GroundingDINO served as the exclusive teacher model. Various state-of-the-art object detection models, including YOLOv8 \cite{yolov8reis2023realtime, yolov8terven2023comprehensive}, YOLOv7 \cite{wang2023yolov7}, and YOLOv5 \cite{Jocher_YOLOv5_by_Ultralytics_2020, yolov8terven2023comprehensive}, were tested with different settings, encompassing parameters like Epoch, Image Size, and Model Variant. This comparative analysis aimed to evaluate detection algorithm performance for camel welfare management in the farm. The experiments offer deployable student model options, augmenting the Fine-Tune Distillation pipeline with GroundingDINO as the teacher model, resulting in a unified deployable framework.

\begin{table}[t]
\centering
\caption{Model hyperparameters}
\label{tab:model_hyperparams}
\resizebox{0.45\textwidth}{!}{%
\begin{tabular}{llllll}
\hline
Model & \multicolumn{1}{c}{Batch} & \multicolumn{1}{c}{Optimizer} & \multicolumn{1}{c}{lr0/lrf} & \multicolumn{1}{c}{Momentum} & \multicolumn{1}{c}{Decay} \\ \hline
YOLOv8 & 16 & SGD & 0.01/0.01 & 0.937 & 0.001 \\ 
YOLOv7 & 16 & SGD & 0.01/0.1 & 0.937 & 0.0005 \\ 
YOLOv5 & 16 & SGD & 0.01/0.01 & 0.937 & 0.0005 \\ \hline
\end{tabular}
}
\end{table}

\begin{figure}[!b]
    \centering
    \includegraphics[width=0.5\textwidth]{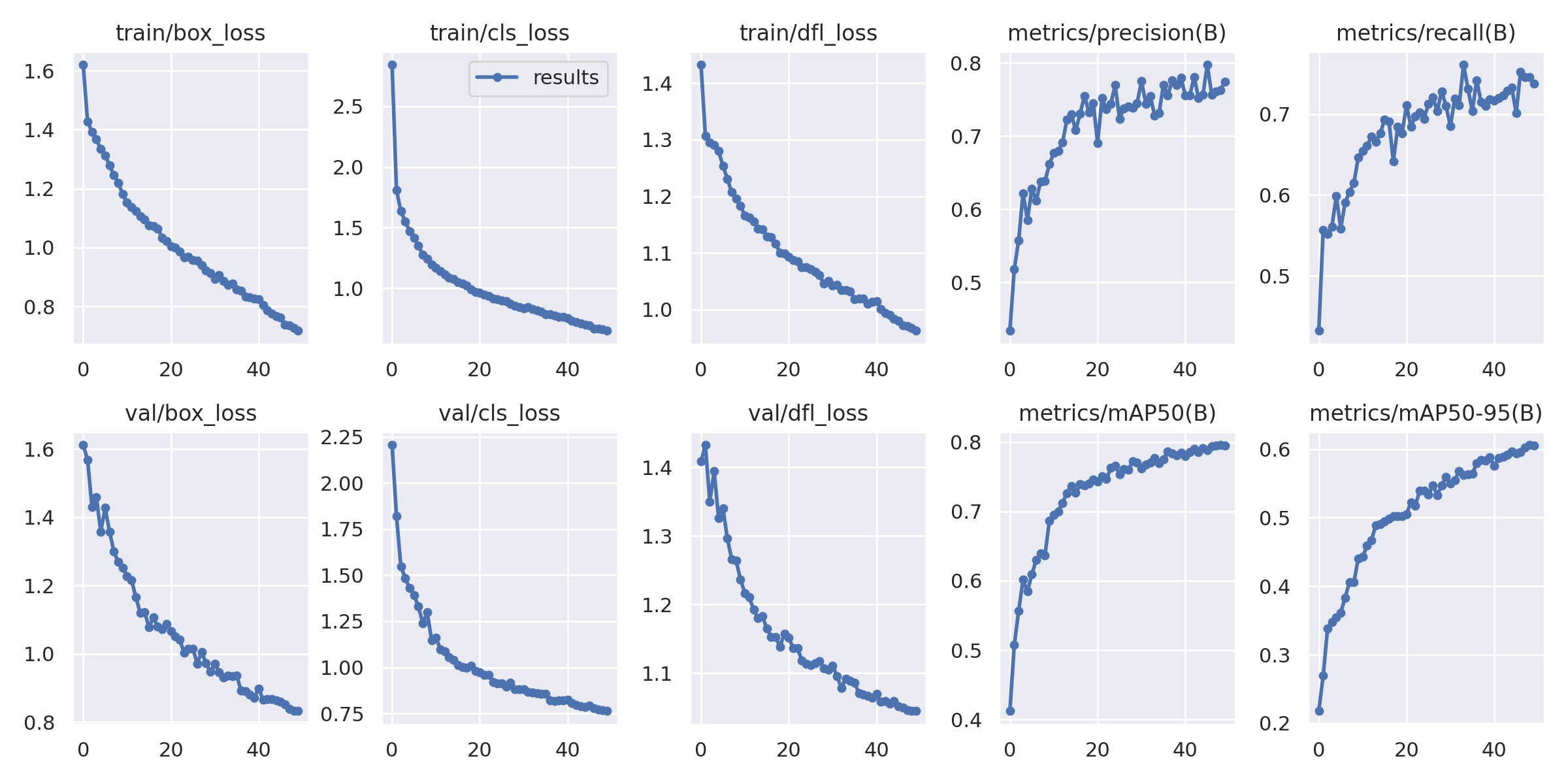}
    \caption{Results of YOLOv8s variant when trained on 50 epochs on Image size of 1024 pixels}
    \label{fig:results_yolov8s}
\end{figure}


\subsection{Models and Metrics}
In the Fine-Tune Distillation approach, following annotations through zero-detection inference, we then fine-tune selected student models (YOLOv8, YOLOv7, and YOLOv5) with the annotated dataset, using various hyperparameters outlined in Table \ref{tab:model_hyperparams}. Across 10 experiments, 8 used different YOLOv8 variants (8s, 8m, and 8n), with 8n being the fastest but less accurate \cite{Yolov8AimAssist}. We also conducted two experiments on YOLOv7 and YOLOv5. Table \ref{tab:model_performance} compares these models' performance across various evaluation metrics and configurations, including experiments with 25 and 50 epochs. Figure \ref{fig:results_yolov8s} presents training results for YOLOv8s with 50 epochs and an image size of 1024 pixels. The experiments employ loss functions such as box loss, class loss, and dual focal loss to evaluate the model's detection capabilities, along with metrics like Average Precision (AP), Recall, mAP50, and mAP50-95 \cite{liu2021loss, dalianis2018evaluation}. These assessments comprehensively gauge the model's performance on the validation dataset.

\[\operatorname{IoU}(A, B)=\frac{|A \cap B|}{|A \cup B|}\]
\[\text{Precision} =\frac{\text { True Positives }}{\text { True Positives }+ \text { False Positives }}\]
\[\text{Recall} =\frac{\text { True Positives }}{\text { True Positives }+ \text { False Negatives }}\]
\[\text{AP}=\sum_n\left(\right. \text{Recall}_n- \text{Recall}\left._{n-1}\right)\cdot \text{Precision}_n\]
\[ \text{mAP}=\frac{\text{AP}_{\text{IoU}=0.5}+\text{AP}_{\text{IoU}=0.55}+\cdots+\text{AP}_{\text{IoU}=0.95}}{\text{k}}
\]

When deploying machine learning models in real-world scenarios like camel farm monitoring, optimizing parameters for edge devices is crucial \cite{lai2018rethinking, shuvo2022efficient, schrom2017evaluation, wenczel2017gaze}. Table \ref{tab:model_props} summarizes these parameters and performance metrics for our experimental models. The number of layers affects model depth and complexity, with more layers capturing intricate patterns but increasing computational costs. The number of parameters represents learnable elements, enhancing complex relationships but increasing memory and computational requirements. Model size is vital for storage-constrained edge devices, favoring compact models that occupy less space while maintaining performance. We conducted additional experiments to evaluate key metrics, including (1) Number of parameters, (2) FLOPS (floating-point operations per second), (3) FPS (frames per second), and (4) Average Precision (AP), which provide insights into model speed. Lower values for parameters and FLOPS indicate efficient resource utilization, while higher FPS and AP scores indicate faster processing and better performance.

\begin{table*}[t]
\centering
\caption{Comparison of the experimented models in terms of their performance using several key metrics. AP=Average Precision across all levels of confidence thresholds. $\text{AP}_{valid 0.50}$=AP at an IoU (Intersection over Union) threshold of 0.50. $\text{AP}_{valid 0.95}$=AP at an IoU (Intersection over Union) threshold of 0.95.}
\label{tab:model_performance}
\resizebox{16cm}{!}{%
\begin{tabular}{llllllllll}
\hline
Model & \multicolumn{1}{c}{Epoch} & \multicolumn{1}{c}{Size} & \multicolumn{1}{c}{AP} & \multicolumn{1}{c}{Recall} & \multicolumn{1}{c}{$\text{AP}_{valid 0.50}$} & \multicolumn{1}{c}{$\text{AP}_{valid 0.95}$} & \multicolumn{1}{c}{$\text{BoxLoss}_{val}$} & \multicolumn{1}{c}{$\text{ClsLoss}_{val}$} & \multicolumn{1}{c}{$\text{DflLoss}_{val}$} \\
\hline
YOLOv8s & 25 & 800 & 0.76233 & 0.7505 & 0.79714 & 0.61101 & 0.81833 & 0.71968 & 0.99883 \\
YOLOv8s & 25 & 512 & 0.75484 & 0.7244 & 0.76765 & 0.57323 & 0.8443 & 0.71241 & 0.9542 \\
YOLOv8m & 25 & 512 & 0.72057 & 0.7232 & 0.76613 & 0.56324 & 0.83586 & 0.72906 & 0.95439 \\
YOLOv8s & 25 & 1024 & 0.78088 & 0.7382 & 0.78985 & 0.61103 & 0.80555 & 0.72885 & 1.0326 \\
YOLOv8s & 25 & 1280 & 0.75429 & 0.7643 & 0.79028 & 0.60044 & 0.87487 & 0.76487 & 1.0431 \\
\rowcolor{Gray}
YOLOv8s & 50 & 1024 & 0.80299 & 0.7294 & 0.79274 & 0.62709 & 0.7502 & 0.72657 & 1.0196 \\
YOLOv8n & 25 & 512 & 0.78159 & 0.6765 & 0.75084 & 0.53241 & 0.9371 & 0.84722 & 0.99122 \\
YOLOv8n & 50 & 1024 & 0.77444 & 0.7376 & 0.79522 & 0.60548 & 0.83274 & 0.76348 & 1.0448 \\
YOLOv7 & 50 & 640 & 0.7696 & 0.6861 & 0.745 & 0.5473 & 0.04332 & 0.02315 & 0.008917 \\
YOLOv5 & 50 & 640 & 0.75475 & 0.7289 & 0.75075 & 0.47346 & 0.02923 & 0.033114 & 0.0051627\\
\hline
\end{tabular}
}
\end{table*}

\begin{figure}[!t]
\centering
    \begin{minipage}{.22\textwidth}
      \centering
      \includegraphics[width=\textwidth]{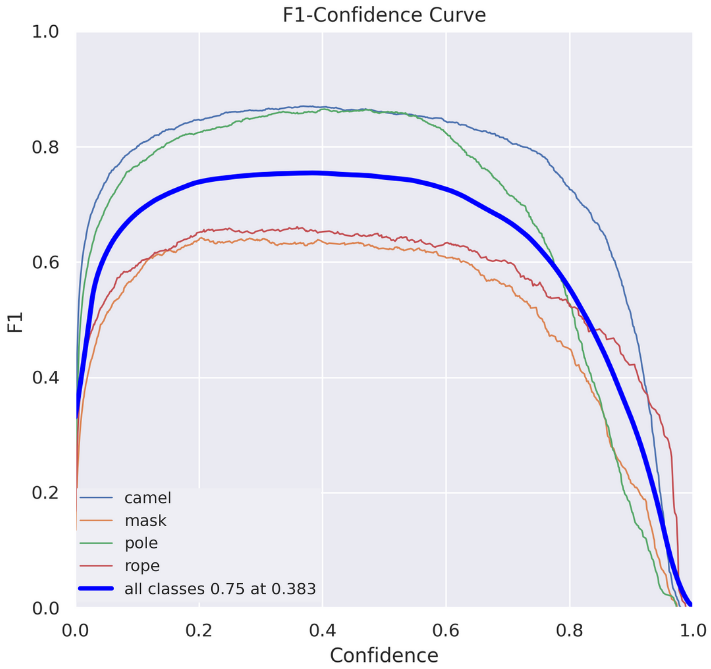}
    \end{minipage}%
    \begin{minipage}{.22\textwidth}
      \centering
      \includegraphics[width=\textwidth]{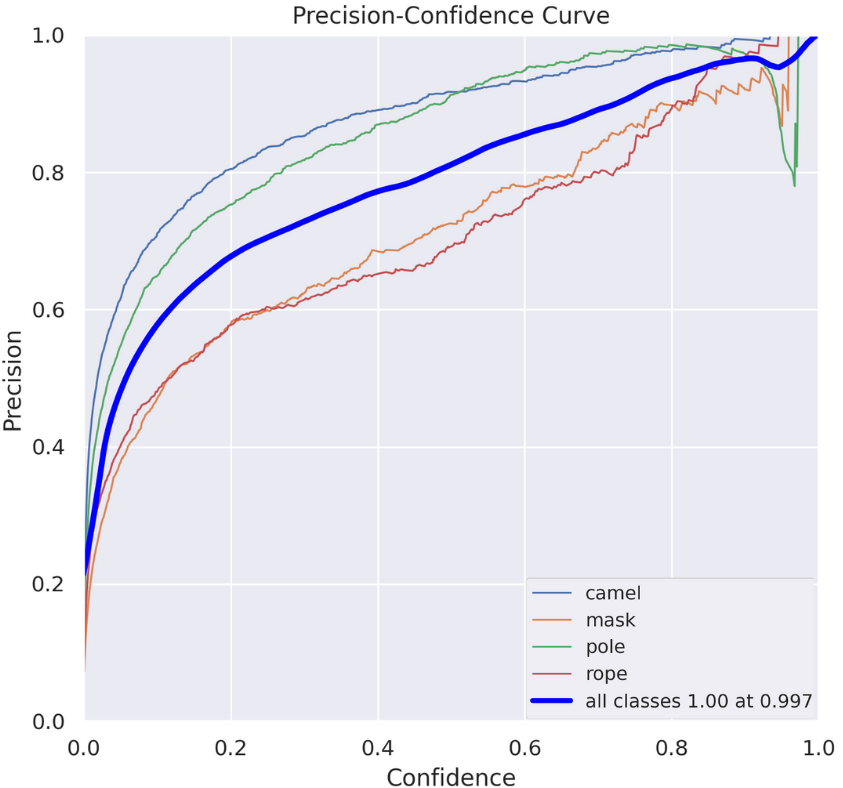}
    \end{minipage}
    \begin{minipage}{.22\textwidth}
      \centering
      \includegraphics[width=\textwidth]{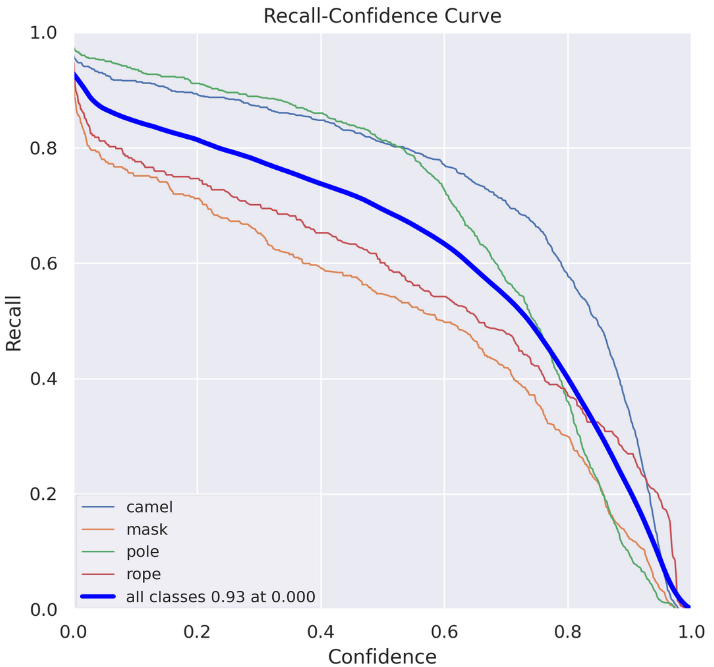}
    \end{minipage}
    \begin{minipage}{.22\textwidth}
      \centering
      \includegraphics[width=\textwidth]{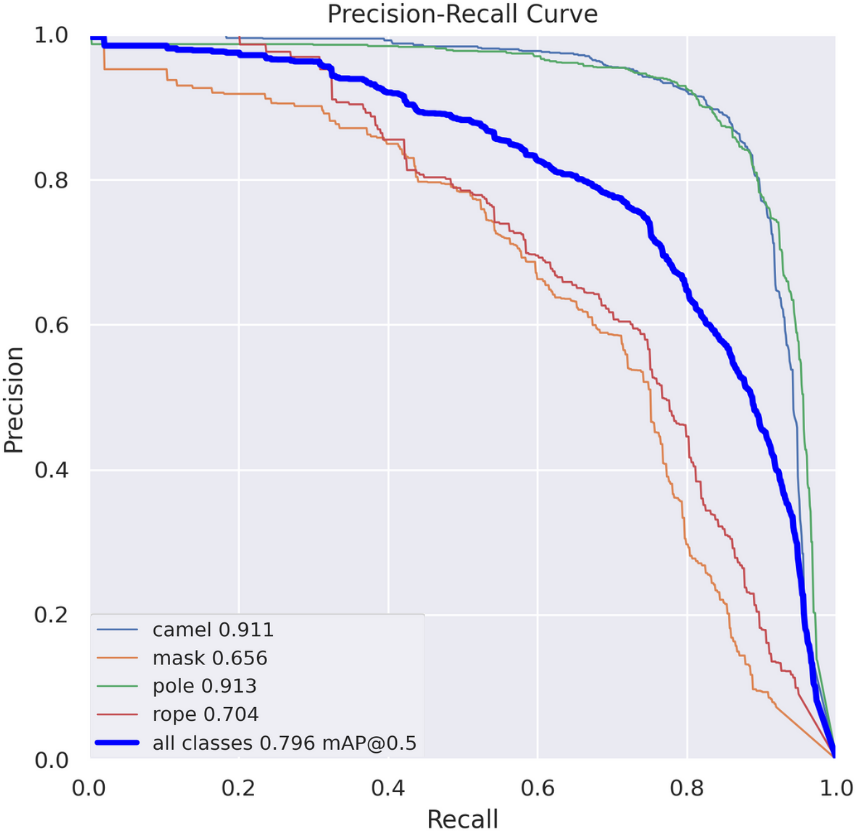}
    \end{minipage}
\caption{Performance of the best performing YOLOv8s model using metrics F1-confidence curve, precision-confidence curve, recall-confidence curve, and precision-recall curve indicating insights into the trade-off between precision, recall, confidence, and F1-score}
\label{fig:four_curves}
\end{figure}

\begin{table}[t]
\centering
\caption{Confusion Matrix of the best performing YOLOv8s student model when trained for 50 epochs on 1024 image size}
\label{tab:confusion_matrix}
\begin{tabular}{llllll}
\hline
\multirow{1}{*}{\makecell{$\text{True} \rightarrow$ \\ $\text{Predicted} \downarrow $}} & \multicolumn{1}{c}{} & \multicolumn{1}{c}{Camel} & \multicolumn{1}{c}{Mask} & \multicolumn{1}{c}{Pole} & \multicolumn{1}{c}{Rope} \\ \cline{2-6} 
 & Camel & \cellcolor{babyblue}{0.88} & 0.01 & 0.01 & 0.01 \\ 
 & Mask & 0.00 & \cellcolor{blizzardblue}{0.63} & 0.00 & 0.04 \\ 
 & Pole & 0.00 & 0.00 & \cellcolor{babyblue}{0.90} & 0.00 \\ 
 & Rope & 0.00 & 0.05 & 0.00 & \cellcolor{blizzardblue}{0.65} \\ \hline 

\end{tabular}
\end{table}

\section{Results}
\subsection{Performance Analysis}
\textbf{Numerical Metrics.} YOLOv8s (50 epochs, Size 1024) performed exceptionally well with an AP of 80.29\%, demonstrating superior detection accuracy. YOLOv8n (25 epochs, Size 512) closely followed with an AP of 78.15\%, showing excellent performance. YOLOv8s (25 epochs, Size 800) achieved an AP of 76.233\%, displaying relatively good accuracy (Table \ref{tab:model_performance}).
Notably, when YOLOv8s was trained with 25 epochs and lower image sizes (512 and 800), its testing performance slightly decreased to approximately 75\%, compared to the 79\% accuracy achieved with an image size of 1024. YOLOv8n, trained with 25 epochs and an image size of 512, and YOLOv8n trained with 50 epochs and an image size of 1024, both achieved strong AP scores of 78.159\% and 77.444\%, respectively.
For Recall, YOLOv8s (25 epochs, Size 1280) excelled with the highest score of 76.43\%, indicating its exceptional ability to capture true positive predictions. YOLOv8s (25 epochs, Size 1024) achieved a recall score of 73.82\%, displaying a relatively high capability. YOLOv8n (50 epochs, Size 1024) also performed well with a recall score of 73.76\%.
In terms of AP50, YOLOv8s (25 epochs, Size 800) achieved the highest score of 79.71\%, signifying accurate object localization with moderate overlap. YOLOv8n (50 epochs, Size 1024) achieved an AP50 score of 79.52\%, indicating high localization accuracy.
Considering all evaluation metrics, YOLOv8s (50 epochs, Size 1024) emerged as the best-performing model, achieving the highest AP\_valid 0.95 score of 62.70\%, outperforming other variants.

{\renewcommand{\arraystretch}{1}
\begin{table*}[t]
\centering
\caption{Comparison of the experimented model properties in terms of computational parameters. Here 1M=$\text{10}^{6}$. 1G=$\text{10}^{9}$. 1Mb=$\text{10}^{3}$. \#=Number. Size=Image size. FLOPs=Floating point operations per second. Weight=Hardisk space required by model weights. FPS=Frames per second during video inference. Inference=Inference time in millisecond. AP=Average Precision}
\label{tab:model_props}
\resizebox{0.75\textwidth}{!}{%
\begin{tabular}{llllllllll}
\hline
Model & \multicolumn{1}{c}{Epoch} & \multicolumn{1}{c}{Size} & \multicolumn{1}{c}{\#Layers} & \multicolumn{1}{c}{\#Params} & \multicolumn{1}{c}{FLOPs} & \multicolumn{1}{c}{Weight} & \multicolumn{1}{c}{FPS} & \multicolumn{1}{c}{Inference (ms)} & \multicolumn{1}{c}{AP} \\ \hline
YOLOv8s & 25 & 800 & 168 & 11.1M & 28.4G & 21.4Mb & 185 & 4.2 & 0.76233 \\
YOLOv8s & 25 & 512 & 168 & 11.1M & 28.4G & 21.4Mb & 370 & 1.5 & 0.75484 \\
YOLOv8m & 25 & 512 & 168 & 25.8M & 78.7G & 49.6Mb & 250 & 2.7 & 0.72057 \\
YOLOv8s & 25 & 1024 & 168 & 11.1M & 28.4G & 21.5Mb & 192 & 3.4 & 0.78088 \\
YOLOv8s & 25 & 1280 & 168 & 11.1M & 28.4G & 21.6Mb & 130 & 5.4 & 0.75429 \\
\rowcolor{Gray} 
YOLOv8s & 50 & 1024 & 168 & 11.1M & 28.4G & 21.5Mb & 182 & 3.4 & 0.80299 \\
YOLOv8n & 25 & 512 & 168 & 3.0M & 8.1G & 5.9Mb & 370 & 1.4 & 0.78159 \\
YOLOv8n & 50 & 1024 & 168 & 3.0M & 8.1G & 6.0Mb & 302 & 1.5 & 0.77444 \\
YOLOv7 & 50 & 640 & 314 & 36.5M & 103.2G & 71.4Mb & 77 & 11.9 & 0.7696 \\
YOLOv5 & 50 & 640 & 182 & 7.2M & 13.1G & 14.6Mb & 156 & 5.1 & 0.75475 \\
\hline
\end{tabular}
}
\end{table*}

\textbf{Graphical Metrics.} Figure \ref{fig:four_curves} visually presents F1-confidence, precision-confidence, recall-confidence, and precision-recall curves curves and their implications.
Our best YOLOv8s model achieves an F1-score of 75\% at around 50\% confidence when considering all classes together. It excels in detecting \textit{Camel} and \textit{Pole} classes with an impressive 86\% F1-score, outperforming \textit{Mask} and \textit{Rope} classes.
In precision-recall analysis, the model's best overall average precision occurs at 80\% precision and 73\% recall. \textit{Camel} and \textit{Pole} classes exhibit about 90\% precision with higher recall, while \textit{Rope} and \textit{Mask} classes show 63\% precision-recall values.
The model performs optimally with 80\% precision for all classes combined. It excels in distinguishing \textit{Camel} and \textit{Pole} classes, as shown in the Confusion Matrix in Table \ref{tab:confusion_matrix}.

After evaluating various models and combinations, YOLOv8s with 50 epochs and an image size of 1024 emerged as the top-performing model, boasting the highest AP, competitive recall, and a strong AP50 score. Therefore, we've chosen YOLOv8s as the final student model for integration into the Fine-Tune Distillation pipeline alongside GroundingDINO, our teacher model. This integration will result in a unified and deployable framework. However, it's important to note that the selection of the most suitable model should consider specific requirements and constraints, such as computational resources and application-specific needs.

\subsection{Computational Analysis}
\textbf{\#Params and FLOPs.} All YOLOv8s variants share a FLOP count of 28.4G, indicating moderate computational complexity regardless of image size and epochs (Table \ref{tab:model_props}). In contrast, YOLOv8m with 25 epochs and a size of 512 demands significantly higher computations with 78.7G FLOPs. YOLOv8n offers lower computational complexity, requiring only 8.1G FLOPs. YOLOv7 is the most computationally demanding model with a FLOP count of 103.2G, while YOLOv5 is relatively lightweight with 13.1G FLOPs.

\textbf{FPS.} YOLOv8s with 25 epochs and a size of 512, along with YOLOv8n with the same configurations, achieved the highest performance, both reaching an FPS of 370, suitable for time-sensitive applications. YOLOv8n with 50 epochs and a size of 1024 closely follows with an FPS of 302. Other models offer FPS ranging from 77 to 250, with YOLOv7 having the lowest FPS of 77.

\textbf{Inference Time.} YOLOv8n and YOLOv8s with a size of 512 demonstrated the fastest inference time of 1.4 milliseconds. YOLOv8n with a size of 1024, as well as YOLOv8s with 25 epochs and 1024 size, achieved slightly higher inference times of 1.5 milliseconds. Other models showed inference times ranging from 2.7 to 11.9 milliseconds, providing options with varying inference time requirements.

\textbf{Precision.} Maintaining a desirable AP is equally crucial. YOLOv8s with 50 epochs and 1024 size emerged as the top performer with an AP of 80.29\%, indicating superior object detection accuracy. YOLOv8s with 25 epochs and 1024 size closely followed with an AP of 78.08\%. These models exhibited high precision in identifying and localizing objects. AP values for other models ranged from 72.05\% to 78.16\%, offering viable options with slightly varying accuracy levels.

The choice of the best model depends on the application's specific requirements. For high FPS and low inference time, YOLOv8s (512) is the top choice. YOLOv8s (1024) offers the highest AP for accuracy-centric applications. YOLOv8n (512) or YOLOv8n (1024) strike a balance between speed and accuracy.
However, for camel farm monitoring, YOLOv8s with 50 epochs and 1024 size emerges as the best model. It balances computational complexity, storage, real-time performance, inference speed, and detection accuracy with 28.4G FLOPs, 21.5Mb weight, 182 FPS, 3.4ms inference time, and an AP of 80.299\%.
Ultimately, the model choice should align with the specific priorities and needs of the application.

\section{Discussion and Limitation}
The Fine-Tune Distillation framework, combining GroundingDINO and YOLOv8, offers several advantages over standalone zero-shot models like GroundingDINO. Firstly, it enables the creation of an initial deployable model version without manual data labeling, reducing time-to-deployment. This is valuable when obtaining labeled data is time-consuming or costly. Secondly, it provides transparency in the training process, allowing visibility into the training data for debugging and data quality improvements. Thirdly, it offers cost savings through automated labeling, reducing human intervention. Additionally, distilled models like YOLOv8 are significantly smaller in size (5Mb to 20Mb) compared to larger models (700Mb to 2.5Gb), making them more practical for storage and deployment. Despite their smaller size, distilled models achieve comparable or superior results (as illustrated in Figure \ref{fig:eg_teacher_vs_stu}), enhancing computational efficiency.

\begin{figure}[t]
\centering
    \begin{minipage}{.5\textwidth}
      \centering
      \includegraphics[width=0.9\textwidth]{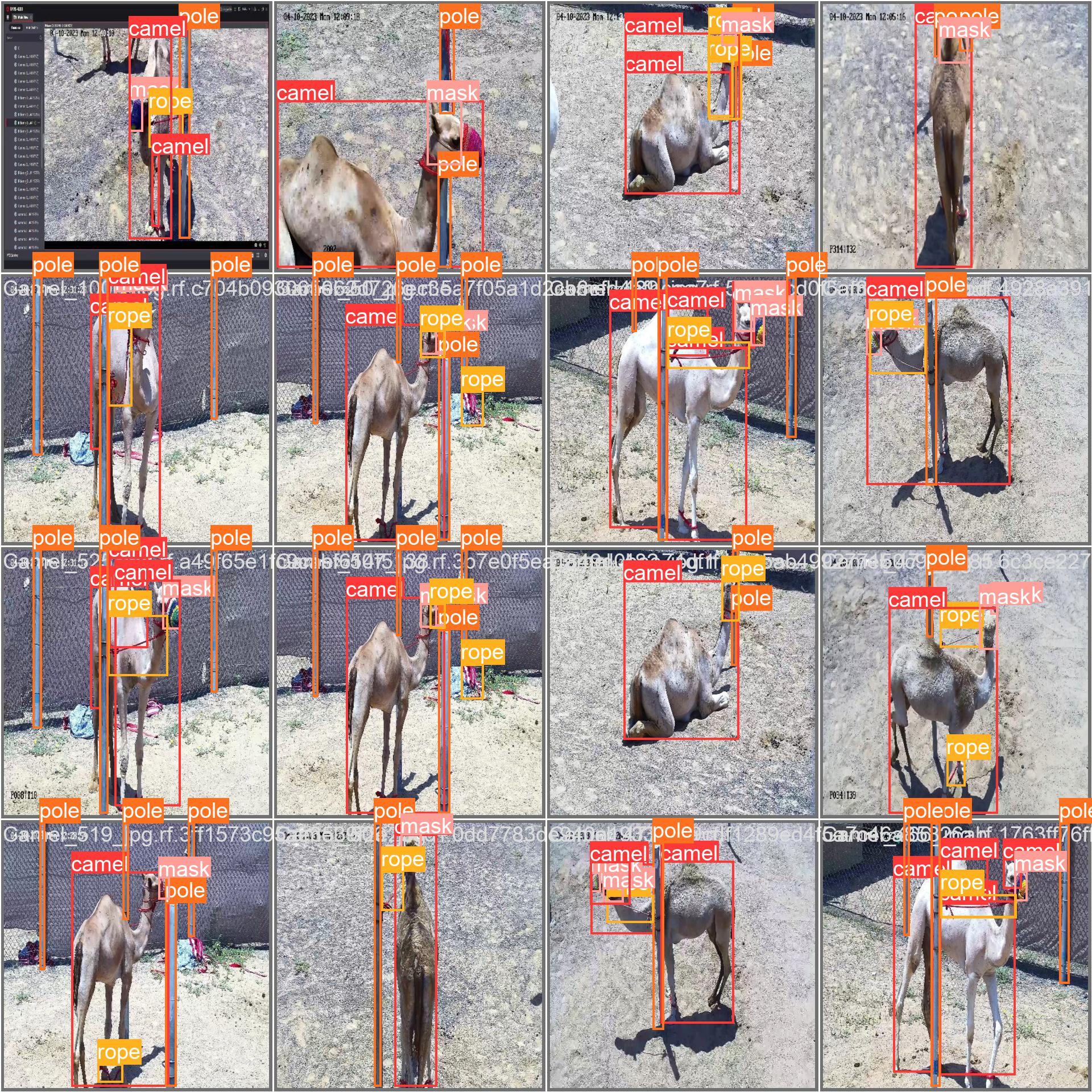}
    \end{minipage}\\
    \begin{minipage}{.5\textwidth}
      \centering
      \includegraphics[width=0.9\textwidth]{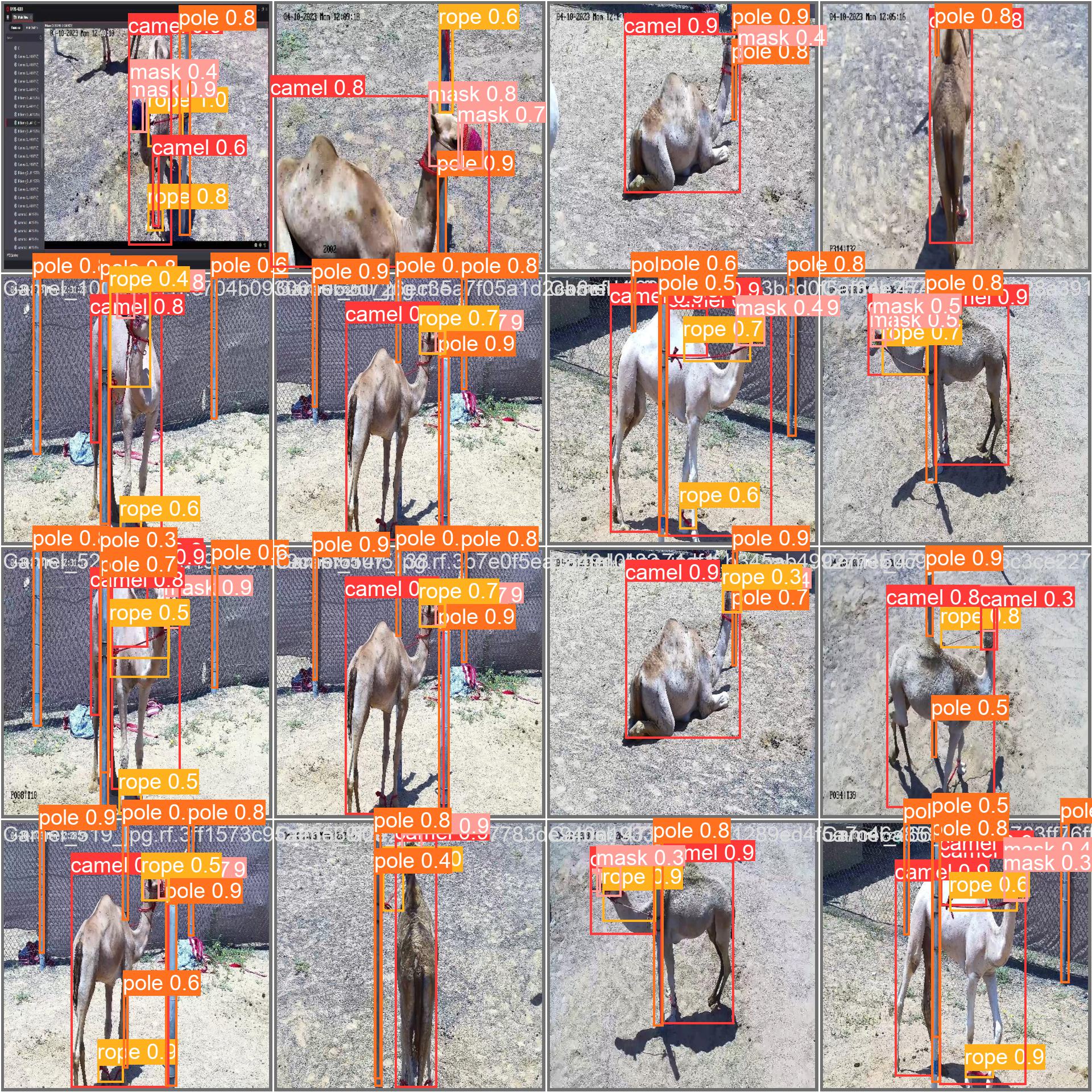}
    \end{minipage}
\caption{The images (left) generated by the Unified Auto-Annotation framework acting as labels v/s Example Predictions (right) by the proposed Fine-tuned distilled YOLOv8 of the labelled images. Here, the \textcolor{red}{red} BB (Bounding Box) denotes class ”camel”, \textcolor{amber}{yellow} denotes ”rope”, \textcolor{cherryblossompink}{pink} denotes ”mask”, and \textcolor{carrotorange}{orange} color BB denotes ”pole”}
\label{fig:eg_teacher_vs_stu}
\end{figure}

\textbf{Limitation.} While the Fine-Tune Distillation framework offers advantages, it has limitations. Teacher models may struggle with identifying all desired classes, especially obscure ones, requiring experimentation or prompt refinement. Some zero-shot models may struggle with classes that have similar contexts in natural language. As new teacher models are developed, improvements are expected. The framework is flexible for future model integration. While it performs well in common domains, it's encouraged to assess suitability for specific use cases. Label accuracy depends on classes; common ones yield better results. For less common classes, quality may be lower. To evaluate model effectiveness, start with a small dataset sample. Full automation of annotation isn't yet feasible; human validation remains vital for optimal outcomes. Nevertheless, models like GroundingDINO and SAM save time when annotating large datasets, allowing focus on result refinement and training more accurate ML models.

\section{Conclusion and Future Work}
Leveraging computer vision capabilities benefits farms by automating tasks and enhancing animal welfare monitoring, like during taming phase. Identifying elements like poles, ropes, masks, and camels enables close monitoring and prompt intervention. In this context, we proposed two key contributions: the Unified Auto-Annotation framework and the Fine-Tune Distillation framework.
The Unified Auto-Annotation approach combines GroundingDINO (GD) and Segment-Anything-Model (SAM) to automatically annotate surveillance video data from the camel farm. The Fine-Tune Distillation framework refines a student model using this annotated dataset through knowledge transfer from a large teacher model. After extensive experimentation with various YOLO variants, YOLOv8s fine-tuned with 50 epochs and an image size of 1024 emerged as the best-performing student model. It achieved the highest average precision (80.3\%) and competitive recall, making it efficient for real-time object detection on our camel dataset. Along with, YOLOv8s with these specifications offers optimal computational complexity, reasonable storage requirements, an FPS of 182, an inference time of 3.4 milliseconds, and the highest AP. This makes it an optimal choice for our use-case, balancing efficiency and accuracy in object detection for camel farm monitoring.

\textbf{Future Work.} In future we aim on enhancing Fine-Tune Distillation by (i) integrating additional sensor modalities like thermal imaging and sound recognition to gather health and stress data, (ii) incorporating real-time feedback to handlers during taming, and (iii) extending monitoring capabilities beyond taming to detect diseases and provide early warnings for health issues. These avenues promise to improve animal welfare, operational efficiency, and overall farm productivity.

\bibliographystyle{unsrt}
\bibliography{references}

\begin{thebibliography}{10}

\bibitem{koger2023quantifying}
Benjamin Koger, Adwait Deshpande, Jeffrey~T Kerby, Jacob~M Graving, Blair~R
  Costelloe, and Iain~D Couzin.
\newblock Quantifying the movement, behaviour and environmental context of
  group-living animals using drones and computer vision.
\newblock {\em Journal of Animal Ecology}, 2023.

\bibitem{tassinari2021computer}
Patrizia Tassinari, Marco Bovo, Stefano Benni, Simone Franzoni, Matteo Poggi,
  Ludovica Maria~Eugenia Mammi, Stefano Mattoccia, Luigi Di~Stefano, Filippo
  Bonora, Alberto Barbaresi, et~al.
\newblock A computer vision approach based on deep learning for the detection
  of dairy cows in free stall barn.
\newblock {\em Computers and Electronics in Agriculture}, 182:106030, 2021.

\bibitem{abd2020review}
Nur Syazarin~Natasha Abd~Aziz, Salwani~Mohd Daud, Rudzidatul~Akmam Dziyauddin,
  Mohamad~Zulkefli Adam, and Azizul Azizan.
\newblock A review on computer vision technology for monitoring poultry
  farm—application, hardware, and software.
\newblock {\em IEEE access}, 9:12431--12445, 2020.

\bibitem{barnard2016quick}
Shanis Barnard, Simone Calderara, Simone Pistocchi, Rita Cucchiara, Michele
  Podaliri-Vulpiani, Stefano Messori, and Nicola Ferri.
\newblock Quick, accurate, smart: 3d computer vision technology helps assessing
  confined animals’ behaviour.
\newblock {\em PloS one}, 11(7):e0158748, 2016.

\bibitem{kirillov2023segment}
Alexander Kirillov, Eric Mintun, Nikhila Ravi, Hanzi Mao, Chloe Rolland, Laura
  Gustafson, Tete Xiao, Spencer Whitehead, Alexander~C. Berg, Wan-Yen Lo, Piotr
  Dollár, and Ross Girshick.
\newblock Segment anything, 2023.

\bibitem{openai2023gpt4}
OpenAI.
\newblock Gpt-4 technical report, 2023.

\bibitem{li2023blip2}
Junnan Li, Dongxu Li, Silvio Savarese, and Steven Hoi.
\newblock Blip-2: Bootstrapping language-image pre-training with frozen image
  encoders and large language models, 2023.

\bibitem{jia2022visual}
Menglin Jia, Luming Tang, Bor-Chun Chen, Claire Cardie, Serge Belongie, Bharath
  Hariharan, and Ser-Nam Lim.
\newblock Visual prompt tuning.
\newblock In {\em European Conference on Computer Vision}, pages 709--727.
  Springer, 2022.

\bibitem{gou2021knowledge}
Jianping Gou, Baosheng Yu, Stephen~J Maybank, and Dacheng Tao.
\newblock Knowledge distillation: A survey.
\newblock {\em International Journal of Computer Vision}, 129:1789--1819, 2021.

\bibitem{wang2023review}
Jiaqi Wang, Zhengliang Liu, Lin Zhao, Zihao Wu, Chong Ma, Sigang Yu, Haixing
  Dai, Qiushi Yang, Yiheng Liu, Songyao Zhang, Enze Shi, Yi~Pan, Tuo Zhang,
  Dajiang Zhu, Xiang Li, Xi~Jiang, Bao Ge, Yixuan Yuan, Dinggang Shen, Tianming
  Liu, and Shu Zhang.
\newblock Review of large vision models and visual prompt engineering, 2023.

\bibitem{liu2023grounding}
Shilong Liu, Zhaoyang Zeng, Tianhe Ren, Feng Li, Hao Zhang, Jie Yang, Chunyuan
  Li, Jianwei Yang, Hang Su, Jun Zhu, et~al.
\newblock Grounding dino: Marrying dino with grounded pre-training for open-set
  object detection.
\newblock {\em arXiv preprint arXiv:2303.05499}, 2023.

\bibitem{Yolov8AimAssist}
Zhang Franklin.
\newblock Yolov8 aim assist.
\newblock \url{https://github.com/Franklin-Zhang0/Yolo-v8-Apex-Aim-assist},
  2023.

\bibitem{falomir2011describing}
Zoe Falomir, Ernesto Jim{\'e}nez-Ruiz, M~Teresa Escrig, and Lled{\'o} Museros.
\newblock Describing images using qualitative models and description logics.
\newblock {\em Spatial Cognition \& Computation}, 11(1):45--74, 2011.

\bibitem{madi2023camel}
Mahmoud Madi, Yasser Basha, Yazan Albadersawi, Fayadh Alenezi, Soliman~Awad
  Mahmoud, Dhafar hamed Abd, Dhiya Al-Jumeily, Wasiq Khan, and Abir~Jaafar
  Hussien.
\newblock Camel detection and monitoring using image processing and iot.
\newblock In {\em 2023 15th International Conference on Developments in
  eSystems Engineering (DeSE)}, pages 305--308. IEEE, 2023.

\bibitem{alnujaidi2023computer}
Khalid Alnujaidi and Ghadah AlHabib.
\newblock Computer vision for a camel-vehicle collision mitigation system.
\newblock {\em arXiv preprint arXiv:2301.09339}, 2023.

\bibitem{alnujaidi2023spot}
Khalid AlNujaidi, Ghada Alhabib, and Abdulaziz AlOdhieb.
\newblock Spot-the-camel: Computer vision for safer roads.
\newblock {\em arXiv preprint arXiv:2304.00757}, 2023.

\bibitem{khojastehkey2019biometric}
Mahdi Khojastehkey, Mohammad Yeganehparast, Alireza Jafari~Arvari, Nader
  Asadzadeh, and Mohammad Khaki.
\newblock Biometric measurement of one-humped camels using machine vision
  technology.
\newblock {\em Journal of Ruminant Research}, 7(1):19--32, 2019.

\bibitem{achour2020image}
Brahim Achour, Malika Belkadi, Idir Filali, Mourad Laghrouche, and Mourad
  Lahdir.
\newblock Image analysis for individual identification and feeding behaviour
  monitoring of dairy cows based on convolutional neural networks (cnn).
\newblock {\em Biosystems Engineering}, 198:31--49, 2020.

\bibitem{shao2020cattle}
Wen Shao, Rei Kawakami, Ryota Yoshihashi, Shaodi You, Hidemichi Kawase, and
  Takeshi Naemura.
\newblock Cattle detection and counting in uav images based on convolutional
  neural networks.
\newblock {\em International Journal of Remote Sensing}, 41(1):31--52, 2020.

\bibitem{cominotte2020automated}
A~Cominotte, AFA Fernandes, JRR Dorea, GJM Rosa, MM~Ladeira, EHCB Van~Cleef,
  GL~Pereira, WA~Baldassini, and OR~Machado Neto.
\newblock Automated computer vision system to predict body weight and average
  daily gain in beef cattle during growing and finishing phases.
\newblock {\em Livestock Science}, 232:103904, 2020.

\bibitem{xu2020automated}
Beibei Xu, Wensheng Wang, Greg Falzon, Paul Kwan, Leifeng Guo, Guipeng Chen,
  Amy Tait, and Derek Schneider.
\newblock Automated cattle counting using mask r-cnn in quadcopter vision
  system.
\newblock {\em Computers and Electronics in Agriculture}, 171:105300, 2020.

\bibitem{wang2019review}
Aichen Wang, Wen Zhang, and Xinhua Wei.
\newblock A review on weed detection using ground-based machine vision and
  image processing techniques.
\newblock {\em Computers and electronics in agriculture}, 158:226--240, 2019.

\bibitem{fan2019brief}
Linwei Fan, Fan Zhang, Hui Fan, and Caiming Zhang.
\newblock Brief review of image denoising techniques.
\newblock {\em Visual Computing for Industry, Biomedicine, and Art},
  2(1):1--12, 2019.

\bibitem{xu2023comprehensive}
Mingle Xu, Sook Yoon, Alvaro Fuentes, and Dong~Sun Park.
\newblock A comprehensive survey of image augmentation techniques for deep
  learning.
\newblock {\em Pattern Recognition}, page 109347, 2023.

\bibitem{shorten2019survey}
Connor Shorten and Taghi~M Khoshgoftaar.
\newblock A survey on image data augmentation for deep learning.
\newblock {\em Journal of big data}, 6(1):1--48, 2019.

\bibitem{newey2018shadow}
Charles~C Newey, Owain~D Jones, and Hannah~M Dee.
\newblock Shadow detection for mobile robots: Features, evaluation, and
  datasets.
\newblock {\em Spatial Cognition \& Computation}, 18(2):115--137, 2018.

\bibitem{zhang2022dino}
Hao Zhang, Feng Li, Shilong Liu, Lei Zhang, Hang Su, Jun Zhu, Lionel~M Ni, and
  Heung-Yeung Shum.
\newblock Dino: Detr with improved denoising anchor boxes for end-to-end object
  detection.
\newblock {\em arXiv preprint arXiv:2203.03605}, 2022.

\bibitem{li2022grounded}
Liunian~Harold Li, Pengchuan Zhang, Haotian Zhang, Jianwei Yang, Chunyuan Li,
  Yiwu Zhong, Lijuan Wang, Lu~Yuan, Lei Zhang, Jenq-Neng Hwang, et~al.
\newblock Grounded language-image pre-training.
\newblock In {\em Proceedings of the IEEE/CVF Conference on Computer Vision and
  Pattern Recognition}, pages 10965--10975, 2022.

\bibitem{li2022dn}
Feng Li, Hao Zhang, Shilong Liu, Jian Guo, Lionel~M Ni, and Lei Zhang.
\newblock Dn-detr: Accelerate detr training by introducing query denoising.
\newblock In {\em Proceedings of the IEEE/CVF Conference on Computer Vision and
  Pattern Recognition}, pages 13619--13627, 2022.

\bibitem{li2023mask}
Feng Li, Hao Zhang, Huaizhe Xu, Shilong Liu, Lei Zhang, Lionel~M Ni, and
  Heung-Yeung Shum.
\newblock Mask dino: Towards a unified transformer-based framework for object
  detection and segmentation.
\newblock In {\em Proceedings of the IEEE/CVF Conference on Computer Vision and
  Pattern Recognition}, pages 3041--3050, 2023.

\bibitem{liu2022dab}
Shilong Liu, Feng Li, Hao Zhang, Xiao Yang, Xianbiao Qi, Hang Su, Jun Zhu, and
  Lei Zhang.
\newblock Dab-detr: Dynamic anchor boxes are better queries for detr.
\newblock {\em arXiv preprint arXiv:2201.12329}, 2022.

\bibitem{birmingham2022multi}
Brandon Birmingham and Adrian Muscat.
\newblock Multi spatial relation detection in images.
\newblock {\em Spatial Cognition \& Computation}, 22(3-4):293--327, 2022.

\bibitem{zhang2022autodistill}
Xiaofan Zhang, Zongwei Zhou, Deming Chen, and Yu~Emma Wang.
\newblock Autodistill: An end-to-end framework to explore and distill
  hardware-efficient language models.
\newblock {\em arXiv preprint arXiv:2201.08539}, 2022.

\bibitem{yolov8terven2023comprehensive}
Juan Terven and Diana Cordova-Esparza.
\newblock A comprehensive review of yolo: From yolov1 to yolov8 and beyond.
\newblock {\em arXiv preprint arXiv:2304.00501}, 2023.

\bibitem{zhou2022detecting}
Xingyi Zhou, Rohit Girdhar, Armand Joulin, Philipp Krähenbühl, and Ishan
  Misra.
\newblock Detecting twenty-thousand classes using image-level supervision,
  2022.

\bibitem{minderer2022simple}
Matthias Minderer, Alexey Gritsenko, Austin Stone, Maxim Neumann, Dirk
  Weissenborn, Alexey Dosovitskiy, Aravindh Mahendran, Anurag Arnab, Mostafa
  Dehghani, Zhuoran Shen, Xiao Wang, Xiaohua Zhai, Thomas Kipf, and Neil
  Houlsby.
\newblock Simple open-vocabulary object detection with vision transformers,
  2022.

\bibitem{zhang2022glipv2}
Haotian Zhang, Pengchuan Zhang, Xiaowei Hu, Yen-Chun Chen, Liunian~Harold Li,
  Xiyang Dai, Lijuan Wang, Lu~Yuan, Jenq-Neng Hwang, and Jianfeng Gao.
\newblock Glipv2: Unifying localization and vision-language understanding,
  2022.

\bibitem{wang2023yolov7}
Chien-Yao Wang, Alexey Bochkovskiy, and Hong-Yuan~Mark Liao.
\newblock Yolov7: Trainable bag-of-freebies sets new state-of-the-art for
  real-time object detectors.
\newblock In {\em Proceedings of the IEEE/CVF Conference on Computer Vision and
  Pattern Recognition}, pages 7464--7475, 2023.

\bibitem{Jocher_YOLOv5_by_Ultralytics_2020}
Glenn Jocher.
\newblock Yolov5 by ultralytics.
\newblock \url{https://github.com/ultralytics/yolov5}, 2020.

\bibitem{YOLONAS_supergradients}
Shay Aharon, {Louis-Dupont}, {Ofri Masad}, Kate Yurkova, {Lotem Fridman},
  {Lkdci}, Eugene Khvedchenya, Ran Rubin, Natan Bagrov, Borys Tymchenko, Tomer
  Keren, Alexander Zhilko, and {Eran-Deci}.
\newblock Super-gradients, 2021.

\bibitem{carion2020endtoend}
Nicolas Carion, Francisco Massa, Gabriel Synnaeve, Nicolas Usunier, Alexander
  Kirillov, and Sergey Zagoruyko.
\newblock End-to-end object detection with transformers, 2020.

\bibitem{cheng2020panopticdeeplab}
Bowen Cheng, Maxwell~D. Collins, Yukun Zhu, Ting Liu, Thomas~S. Huang, Hartwig
  Adam, and Liang-Chieh Chen.
\newblock Panoptic-deeplab: A simple, strong, and fast baseline for bottom-up
  panoptic segmentation, 2020.

\bibitem{yolov8reis2023realtime}
Dillon Reis, Jordan Kupec, Jacqueline Hong, and Ahmad Daoudi.
\newblock Real-time flying object detection with yolov8.
\newblock {\em arXiv preprint arXiv:2305.09972}, 2023.

\bibitem{liu2021loss}
Peidong Liu, Gengwei Zhang, Bochao Wang, Hang Xu, Xiaodan Liang, Yong Jiang,
  and Zhenguo Li.
\newblock Loss function discovery for object detection via
  convergence-simulation driven search, 2021.

\bibitem{dalianis2018evaluation}
Hercules Dalianis and Hercules Dalianis.
\newblock Evaluation metrics and evaluation.
\newblock {\em Clinical text mining: secondary use of electronic patient
  records}, pages 45--53, 2018.

\bibitem{lai2018rethinking}
Liangzhen Lai and Naveen Suda.
\newblock Rethinking machine learning development and deployment for edge
  devices.
\newblock {\em arXiv preprint arXiv:1806.07846}, 2018.

\bibitem{shuvo2022efficient}
Md~Maruf~Hossain Shuvo, Syed~Kamrul Islam, Jianlin Cheng, and Bashir~I Morshed.
\newblock Efficient acceleration of deep learning inference on
  resource-constrained edge devices: A review.
\newblock {\em Proceedings of the IEEE}, 2022.

\bibitem{schrom2017evaluation}
Helmut Schrom-Feiertag, Volker Settgast, and Stefan Seer.
\newblock Evaluation of indoor guidance systems using eye tracking in an
  immersive virtual environment.
\newblock {\em Spatial Cognition \& Computation}, 17(1-2):163--183, 2017.

\bibitem{wenczel2017gaze}
Flora Wenczel, Lisa Hepperle, and Rul von St{\"u}lpnagel.
\newblock Gaze behavior during incidental and intentional navigation in an
  outdoor environment.
\newblock {\em Spatial Cognition \& Computation}, 17(1-2):121--142, 2017.

\end{thebibliography}

\clearpage

\onecolumn
\section*{
\centering
\Large{APPENDIX\\ Domain Adaptable Fine-Tune Distillation Framework For Advancing Farm Surveillance}
}

\begin{table*}[h]
  \centering
  \caption{Summary of related works that signifies the utilization of Computer Vision for Detection, Monitoring, and Farm Management in Camels and Other Livestock}
  \label{tab:related_works}
  \adjustbox{max width=\textwidth}{
  \begin{tabular}{lllll}
    \hline
    Refs. & Contribution & Algo. Used & Data Source & Cam. Perspective \\
    \hline
    \cite{madi2023camel} & Detection, Monitoring & - & GPS \& Drone & 360\textdegree \text{ View} \\
    \cite{alnujaidi2023spot} & Detection on roads & YOLOv8 & Road CAMs & Multiple side angles \\
    \cite{khojastehkey2019biometric} & Body dimension estimation & ANN & Constant distance CAMs & Multiple angles \\
    \cite{achour2020image} & Identification, Behaviour monitoring & CNN & Dairy Cow farm, UAV & UAV \\
    \cite{shao2020cattle} & Detection, Counting & CNN & Aerial UAV & Top, Side View \\
    \cite{cominotte2020automated} & Weight gain estimation & Regression variants, ANN & Cattle farm & Top View \\
    \cite{xu2020automated} & Counting & Mask R-CNN & Quadcopter & Top-view \\
    \hline
  \end{tabular}
  }    
\end{table*}

{\renewcommand{\arraystretch}{1.}
\begin{table}[t]
    \centering
    \caption{Preprocessing and Augmentation Steps implemented in the Data preprocessing phases}
    \label{tab:preprocessing}
    \adjustbox{max width=0.75\textwidth}{
        \begin{tabular}{ll|ll}
        \hline
        \multicolumn{2}{c|}{{Preprocessing Steps}} & \multicolumn{2}{c}{{Augmentation Steps}} \\ \hline
        {Factor} & {Value} & {Factor} & {Value} \\ \hline
        {Brightness} & 1.2 & {Crop} & 0.35 max zoom \\ 
        {Contrast} & 1.5 & {Grayscale} & 20\% x 1502 imgs \\ 
        {Mean} & (0.5, 0.5, 0.5) & {Output/Train Img} & 2 \\ 
        {Standard Deviation} & (0.5, 0.5, 0.5) & {-} & {-} \\ 
        {Noise Kernel} & 3x3 & {-} & {-} \\ \hline
        \end{tabular}
        }
\end{table}

\begin{figure*}[h]
\centering
  \includegraphics[width=0.75\textwidth]{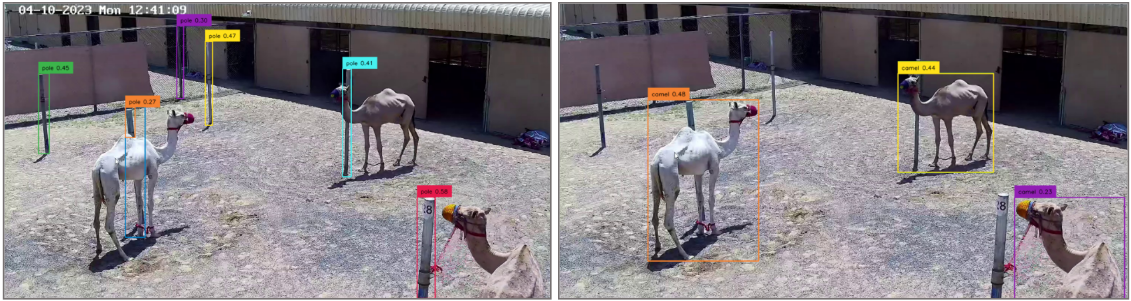}
\caption{Zero-Shot Inference utilizing GroundingDINO on our Camel dataset. When the input to GroundingDINO is prompted with a class of interest: "pole" (left) V/s When a more descriptive input, specific to a use case, is prompted as "camel next to pole" (right). (In left, all the bounding boxes are denoting the class "pole", whereas in right, bounding boxes are denoting the class "camel")}
\label{fig:zero_infer_generic_camel}
\end{figure*}

\begin{figure*}[h]
    \centering
    \includegraphics[width=0.75\textwidth]{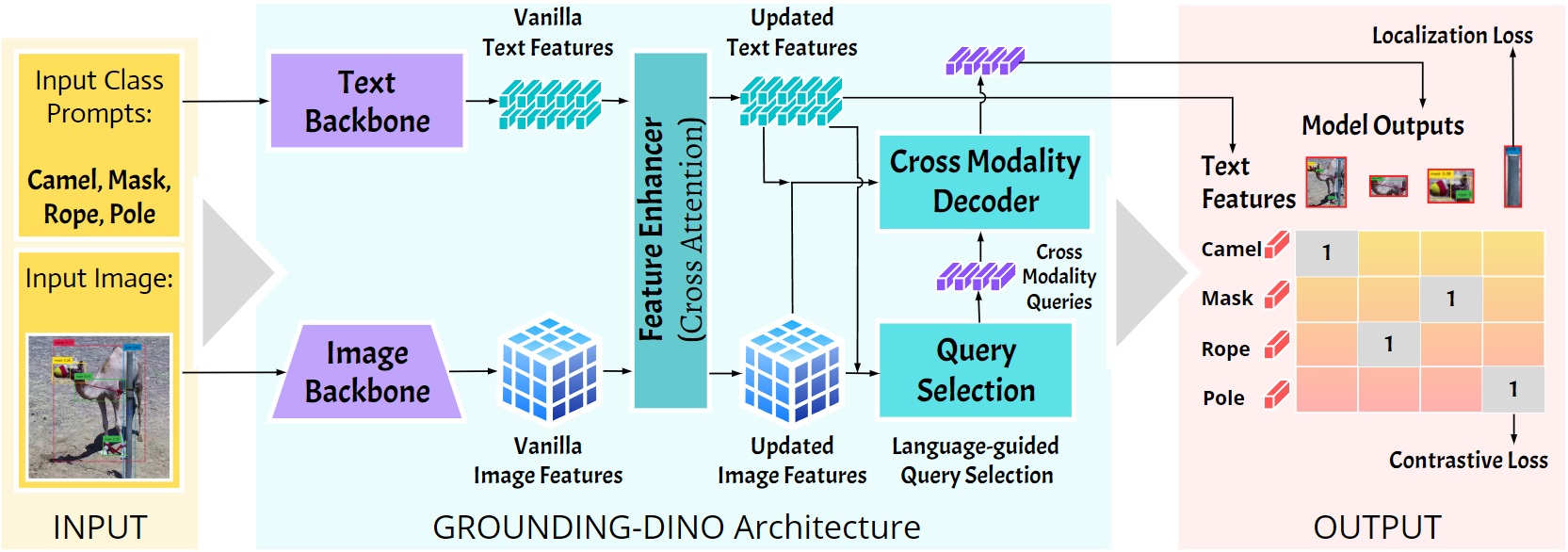}
    \caption{Technical workflow of GroundingDINO for zero-shot detection on the camel farm dataset. Camel images and class prompts are inputted to the Large vision model like GroundingDINO that then performs zero-shot detection to output bounding boxes respective to prompted classes}
    \label{fig:method1_zero_infer}
\end{figure*}

\begin{figure*}[t]
    \centering
    \includegraphics[width=0.75\textwidth]{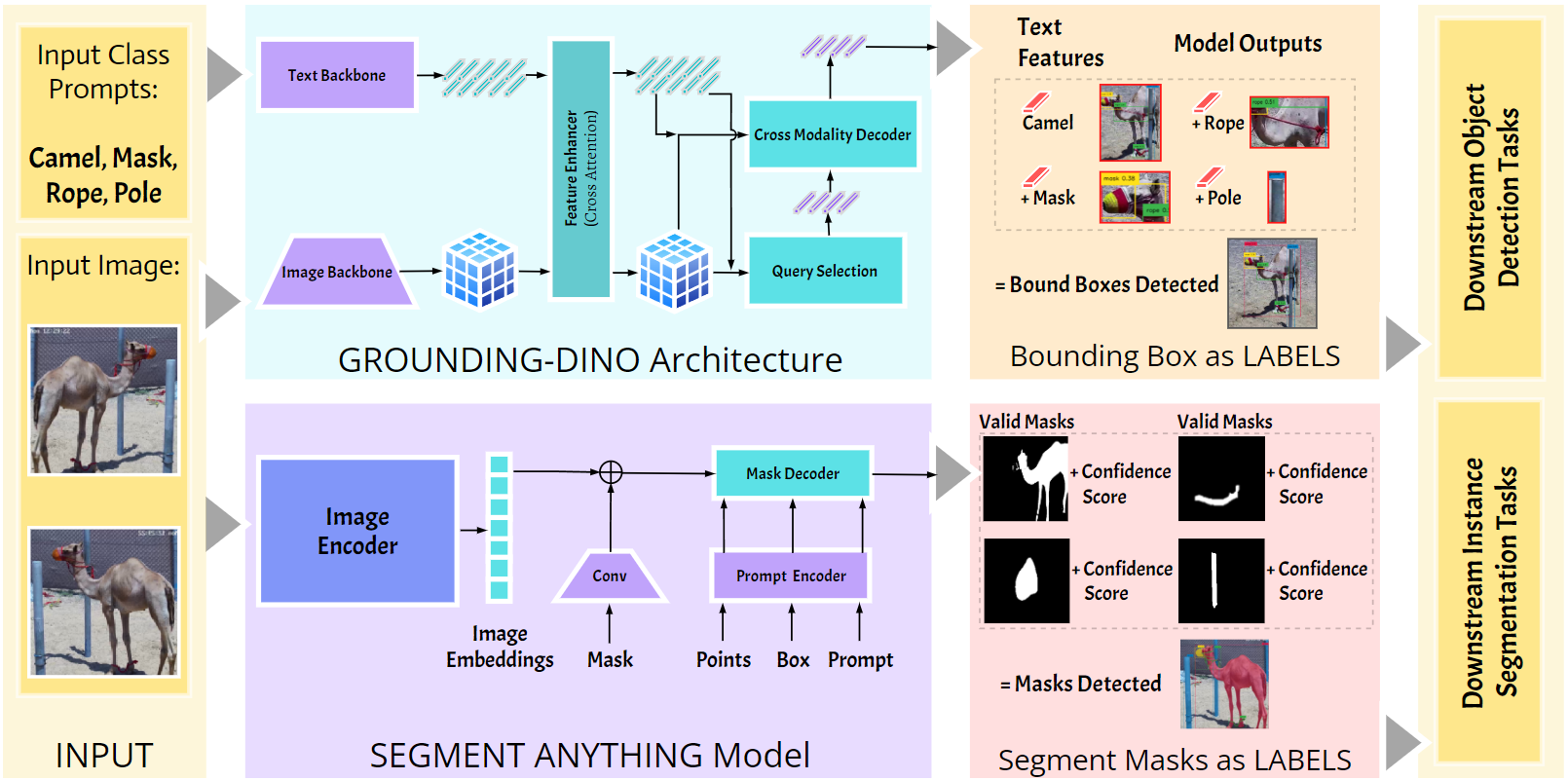}
    \caption{Workflow of the unified auto-annotation framework. Images and class prompts are inputted to GroundingDINO that performs zero-shot detection to output bounding boxes respective to prompted classes. These bounding boxes are then inputted to SAM that outputs segmentation masks respective to the bounding boxes. These bounding boxes and masks are saved and act as labels which can be implemented towards downstream detection or/and segmentation tasks}
    \label{fig:method2_auto_annotation}
\end{figure*}

\begin{algorithm*}[h]
\caption{Pseudocode for the Unified Auto-Annotation Process using GroundingDINO and SAM.\\ \textbf{Input:} Dataset of images with class prompts of interest;
\textbf{Output:} Labels as bounding boxes and segmentation masks;}\label{alg:annotation-process}
\begin{algorithmic}[1]
\State $dataset \gets \text{load\_dataset()}$ \Comment{Load dataset of images}
\State $class\_prompts \gets$ $\text{Input}\{C_1, C_2, ..., C_m\}$ \Comment{Define class prompts for desired classes}
\State $\mathcal{F} \gets$ Pre-trained GroundingDINO model
\State $\mathcal{H} \gets$ Pre-trained SAM model
\State $detection\_dataset \gets [\:]$ \Comment{Initialize empty list for object detection annotations}
\State $segmentation\_dataset \gets [\:]$ \Comment{Initialize empty list for instance segmentation annotations}

\For{each $image$ in $dataset$}
\State $box\_prediction \gets \mathcal{F}(image, class\_prompts)$ \Comment{Perform inference on GroundingDINO}
\State $bb\_annotation \gets (image, box\_prediction)$ \Comment{BB = Boundary Box}
\State $detection\_dataset \gets detection\_dataset\oplus{bb\_annotation}$ \Comment{$\oplus$ = Concatenate}
\EndFor

\For{each $(image,bb\_annotation)$ in $detection\_dataset$}
\State $mask\_prediction \gets \mathcal{H}(image,bb\_annotation)$ \Comment{Perform inference on SAM using BB}
\State $mask\_annotation \gets (image, mask\_prediction)$
\State $segmentation\_dataset \gets segmentation\_dataset\oplus{mask\_annotation}$ \Comment{$\oplus$ = Concatenate}
\EndFor  
\State \textbf{return} $detection\_dataset$, $segmentation\_dataset$ \Comment{Return Detection, Segmentation datasets}
\end{algorithmic}
\end{algorithm*}

\begin{algorithm*}[h]
\caption{Pseudocode for Working of Fine-Tune Distillation. \\ \textbf{Input:} Dataset of images with classes of interest;
\textbf{Output:} Deployable model for real-time applications;}\label{alg:fine-tuning}
\begin{algorithmic}[1]
\State $dataset \gets \text{load\_dataset()}$ \Comment{Load dataset of images}
\State $class\_prompts \gets$ $\text{Input}\{C_1, C_2, ..., C_m\}$ \Comment{Define class prompts for desired classes}
\State $\mathcal{F} \gets$ Pre-trained Teacher model architecture \Comment{Initialize GroundingDINO}
\State $annotated\_dataset \gets [\:]$
\For{each $image$ in $dataset$}
    \State $box\_prediction \gets \mathcal{F}(image, class\_prompts)$ \Comment{Inferring Teacher model prediction}
    \State $annotated\_sample \gets (image, box\_prediction)$
    \State $annotated\_dataset \gets annotated\_dataset\oplus{annotated\_sample}$ \Comment{$\oplus$ = Concatenate}
\EndFor

\State $\mathcal{G} \gets$ Student model architecture \Comment{Initialize YOLOv8}
\State $annotated\_images \gets \text{Load}(annotated\_dataset)$ \Comment{Load annotated images from Step 9}
\State $hyperparams \gets \{\text{num\_epochs, batch\_size, lr, ..., optimizer}\}$ \Comment{Set fine-tuning hyperparameters}
\State $num\_epochs \gets hyperparams.\text{num\_epochs}$
\State $batch\_size \gets hyperparams.\text{batch\_size}$
\For{$epoch = 1$ to $num\_epochs$} \Comment{Fine-tuning loop: Training student model}
    \State $shuffled\_images \gets \text{shuffle}(annotated\_images)$
    \State $mini\_batches \gets \text{split}(shuffled\_images, batch\_size)$ \Comment{Split $shuffled\_images$ into $batch\_size$}
    \For{each mini-batch in the $mini\_batches$}
        \State $(images, annotations) \gets mini\_batch$ \Comment{Extract $images$ and $annotations$ from mini-batch}
        \State $predictions \gets$ $\mathcal{G}.\text{forward}(images)$
        \State $loss \gets$ $\text{calculate\_loss}(predictions, annotations)$
        \State $\mathcal{G}.\text{backward}(loss)$
        \State $updated\_loss +\gets loss$ \Comment{Update the loss value in the $updated\_loss$ variable}
    \EndFor
\EndFor

\State \textbf{return} fine-tuned $\mathcal{G}$ \Comment{Return small student model, ready for deployment}
\end{algorithmic}
\end{algorithm*}

\end{document}